\documentclass[journal,onecolumn]{IEEEtran}
\usepackage{amsmath}
\usepackage{color}
\usepackage{cite}
\usepackage{amsmath}
\usepackage{mathtools}
\usepackage{amsfonts}
\usepackage{hyperref}
\usepackage{xcolor} 
\usepackage{cases}
\usepackage{multirow}
\usepackage{booktabs}
\usepackage[caption=false]{subfig}
\usepackage{diagbox}
\usepackage{nicematrix}
\usepackage[bottom,symbol]{footmisc}
\usepackage[T1]{fontenc}
\usepackage[utf8]{inputenc}
\usepackage{babel}
\usepackage[font=small,labelfont=bf]{caption}
\usepackage{tabularx}
\usepackage[export]{adjustbox}
\usepackage{caption}
\usepackage{booktabs}
\usepackage{varwidth}
\usepackage[ruled, lined, linesnumbered, commentsnumbered, longend]{algorithm2e}
\usepackage{mathtools}
\usepackage[caption=false]{subfig}
\usepackage{commath}
\usepackage{enumitem}

\definecolor{m}{RGB}{240,0,240}
\newcommand{\transp}{^T}
\newcommand*\xbar[1]{%
  \hbox{%
    \vbox{%
      \hrule height 0.5pt 
      \kern0.5ex
      \hbox{%
        \kern-0.1em
        \ensuremath{#1}%
        \kern-0.1em
      }%
    }%
  }%
}

\SetCommentSty{mycommfont}
\let\norm\undefined 
\DeclarePairedDelimiter\norm{\lVert}{\rVert}
\setenumerate[1]{label={(\arabic*)}} 
\ifCLASSINFOpdf
\else
\fi

\begin{document}

\title{Fast Scalable Image Restoration using Total Variation Priors and Expectation Propagation}

\author{Dan Yao,
        Stephen~McLaughlin~\IEEEmembership{Fellow,~IEEE,}
        and~Yoann Altmann,~\IEEEmembership{Member,~IEEE}
\thanks{D. Yao, S. McLaughlin and Y. Altmann are with the School of Engineering and Physical Sciences, Heriot-Watt University, EH14 4AS, Edinburgh, United Kingdom, e-mail: Y.Altmann@hw.ac.uk.}
\thanks{This work was supported by the UK Royal Academy of Engineering under the Research Fellowship Scheme (RF201617/16/31) and by the Engineering and Physical Sciences Research Council of the UK (EPSRC) Grant number EP/S000631/1 and the UK MOD University Defence Research Collaboration (UDRC) in Signal Processing.}}


\maketitle

\begin{abstract}

This paper presents a scalable approximate Bayesian method for image restoration using total variation (TV) priors. In contrast to most optimization methods based on maximum a posteriori estimation, we use the expectation propagation (EP) framework to approximate minimum mean squared error (MMSE) estimators and marginal (pixel-wise) variances, without resorting to Monte Carlo sampling. For the classical anisotropic TV-based prior, we also propose an iterative scheme to automatically adjust the regularization parameter via expectation-maximization (EM). Using Gaussian approximating densities with diagonal covariance matrices, the resulting method allows highly parallelizable steps and can scale to large images for denoising, deconvolution and compressive sensing (CS) problems. The simulation results illustrate that such EP methods can provide a posteriori estimates on par with those obtained via sampling methods but at a fraction of the computational cost. Moreover, EP does not exhibit strong underestimation of  posteriori variances, in contrast to variational Bayes alternatives. 
\end{abstract}

\begin{IEEEkeywords}
Variational inference, image restoration, Expectation Propagation (EP), Expectation Maximization (EM), hyperparameter estimation.
\end{IEEEkeywords}

\IEEEpeerreviewmaketitle

\section{Introduction}
\IEEEPARstart{I}{mage} 
restoration, the recovery of an unknown true image from its degraded measurement, is a fundamental problem in modern image processing \cite{DigitalImage, gunturk2012image}. It has found numerous applications in remote sensing \cite{bioucas2013hyperspectral}, medical imaging \cite{webb1985constrained}, astronomical imaging \cite{molina2001image}, defense and security \cite{bourlai2011restoring}, to name a few. To cope with the ill-posed nature of image restoration, a large number of image processing algorithms have been proposed using the Bayesian formalism or penalty-based formulations \cite{Karl1987illpose,tikhonov1963solution, bioucas2006adaptive}.

Bayesian image restoration often relies on different prior models for the unknown image of interest. Among the broad variety of existing image prior models, this work focuses on priors promoting small image gradients, centered around the classical Total Variation (TV) prior introduced in \cite{rudin1992nonlinear}. Although more advanced priors, such as priors built using convolution neural network \cite{ulyanov2018deep,mataev2019deepred, vaksman2020lidia}, can provide better image estimates, the TV prior model still offers practical advantages. Being a local (Markovian) model, the prior can be evaluated efficiently, i.e., without large matrix multiplication as when using wavelet/dictionary-based priors. When defined using the $\ell_1$ or $\ell_2$ norm of image gradients, the TV prior is log-concave. It also requires only a reduced number of hyperparameters to be tuned and does not require to be trained using external images. This makes TV-based image restoration methods appealing for applications where fast and flexible restoration methods and reliable estimates are preferred over high-quality image estimates. 

While most image restoration methods aim at maximizing a penalised likelihood or a posterior distribution, they often only provide point estimates and limited tools to assess the uncertainties associated with the estimated images \cite{robert2007bayesian, repetti2019scalable, pereyra2017maximum}. The classical approach to uncertainty quantification (UQ) a posteriori when exact computation is not possible remains Monte Carlo sampling. For high-dimensional images, efficient Markov chain Monte Carlo (MCMC) methods have been proposed over the last few years \cite{pakman2014exact,bubeck2015finite,durmus2018efficient,lehec2021langevin}, in particular for log-concave but non-smooth posteriors \cite{durmus2018efficient,lehec2021langevin}, allowing shorter chains to be used, with lower per iteration cost, e.g., using variable splitting \cite{vono2020asymptotically}.  However, there is still a widely held perception that scaling MCMC to modern high-dimensional problems is not (yet) feasible for fast inference \cite{rajaratnam2015mcmc}.
The most popular alternative to sampling is variational approaches \cite{bishop2006pattern, Blei_2017}, which aim to approximate the posterior distribution of interest by a more tractable distribution whose moments are easier to compute. Variational Bayes (VB) methods \cite{attias2000variational} are a classical family of tools used in such cases, yet they can be difficult to implement when likelihoods and priors are not conjugate. While VB methods can efficiently approximate (marginal) posterior means, they tend to  underestimate marginal posterior variances and these estimated quantities should be handled carefully within any subsequent decision-making process. 

Expectation propagation (EP) \cite{MinkaEP,wainwright2008graphical} is another variational alternative to sampling, which has become a popular approximation method for inference in large scale statistical models. This family of algorithms provides efficient solutions to perform approximate Bayesian inference, and has recently been applied to solve high-dimensional imaging problems \cite{ko2016expectation,yao2021joint, braunstein2020compressed, muntoni2019nonconvex}. One of the key differences between EP and VB is the form of the divergences to be minimized, which leads to different approximate probability distributions. By choosing approximations within the exponential family \cite{seeger2005expectation},  EP can be applied to wider sets of problems compared to VB. In contrast to VB, EP tends to overestimate marginal variances and can thus be used more reliably for uncertainty upper bounds in decision-making \cite{jylanki2011robust, lakshminarayanan2016simple}.  

EP has recently been used in combination with gradient-based priors for tomographic image reconstruction \cite{muntoni2019nonconvex}.  However, the resulting EP algorithm has several limitations. Firstly, for an $N$ dimensional image vector, it requires ${\mathcal O}(N^4)$ arithmetic operations per iteration to update the variance of the approximating factor and it is limited to a sequential update scheme, which prevents scalable inference for high-dimensional imaging problems. 
In this work, a set of fast and scalable EP algorithms, which only capture marginal (pixel-wise) variances a posteriori, are proposed to solve high-dimensional image restoration problems. It allows the use of a reduced number of sequential EP updates, where furthermore most steps can be implemented in a parallel fashion. This paper first considers the $\ell_1$-TV prior, which allows the comparison with existing MCMC and VB methods. Furthermore, as in \cite{muntoni2019nonconvex}, we also show that our method can be used with more aggressive (non-convex) gradient-based priors, e.g. spike-and-slab prior, which tends to perform better than $\ell_1$-TV prior in homogeneous regions denoising while preserving sharp edges. Finally, for $\ell_1$-TV prior we illustrate how  the  proposed EP method can be embedded within larger inference schemes. More precisely, we introduce auxiliary variables, without increasing significantly the computational footprint of EP, to allow for the hyperparameter estimation via an Expectation Maximization (EM)-like procedure.

The main contributions of this paper are twofold:
\begin{itemize}
  \item A set of new scalable EP algorithms with convex and non-convex gradient-based priors are proposed to solve high-dimensional image restoration problems, including denoising, non-blind deconvolution, and compressive sensing (CS) reconstruction. These algorithms benefit from closed-form expressions for most updates. 
  \item We illustrate how the EP method can be used efficiently within larger inference schemes, e.g., to estimate the hyperparameter of $\ell_1$-TV prior, despite only capturing marginal variances a posteriori.
\end{itemize}

The reminder of this paper is organized as follows. Section \ref{Sec: Exact Bayesian model} presents an exact Bayesian model with three gradient-based priors considered for image restoration. Section \ref{Sec: Proposed EP algorithm with TV priors} proposes the EP algorithms  to perform approximate Bayesian inference. Hyperparameter estimation is discussed in Section \ref{Sec: EP-EM for hyperparameter estimation of TV-L1 prior}. Section \ref{Sec: Experimental Results} evaluates the performance of the proposed EP algorithms on high-dimensional image restoration problems. Conclusions and further work are finally reported in Section \ref{Sec: Conclusion}.

\section{Bayesian model for image restoration}
\label{Sec: Exact Bayesian model}
The image restoration problem investigated in this work consists of recovering, from a set of observations $\boldsymbol y\in{\mathbb R}^{M}$, an unknown image $\boldsymbol x \in {\mathbb R}^{N}$ which has been linearly transformed by a known degradation operator $\textbf{H} \in {\mathbb R}^{M\times N}$ ($M \leq N$) and corrupted by additive noise. More precisely, using the Bayesian formalism, we aim at providing a point estimate for $\boldsymbol x$ (e.g., posterior mean) and pixel-wise \textit{a posteriori} uncertainty measures.

\subsection{Likelihood}
The observations in  $\boldsymbol y = [y_1,\dots,y_M]^T$ are assumed to be corrupted by independently and identically distributed (i.i.d.) zero-mean Gaussian noise with variance $\xi$ and the mean of $\boldsymbol y$ is $\textbf{H}\boldsymbol x$. The resulting likelihood $f_y(\boldsymbol y|{\mathbf H}\boldsymbol x)$ can be expressed as
\begin{equation}
f_y(\boldsymbol y|\textbf{H}\boldsymbol x) = \prod\limits_{m=1}^M {\mathcal N}(y_m;\boldsymbol h_m\boldsymbol x,\xi),
\label{Eq: likelihood}
\end{equation}
where $\{\boldsymbol h_m\}_{m=1,\dots,M}$ are the rows of $\textbf{H}$. Different structures of matrix $\textbf{H}$ are considered in this work: (\romannumeral 1) $\textbf{H}= \mathbf{I}_N$ is the identity matrix (denoising problem), (\romannumeral 2) $\textbf{H}\in \mathbb{R}^{N\times N}$ is a convolution matrix (image deconvolution problem), (\romannumeral 3) $\textbf{H}\in \mathbb{R}^{M\times N}$ ($M\ll N$) is a sensing matrix in CS, such as a random matrix with Gaussian i.i.d. entries or a subsampled Paley-ordered 2D Hadamard matrix \cite{moshtaghpour2020close}. 

\subsection{Image gradient-based priors}
\label{Subsec: TV_priors}

\begin{figure}[ht!]
\centerline{\includegraphics[width=0.4\textwidth]{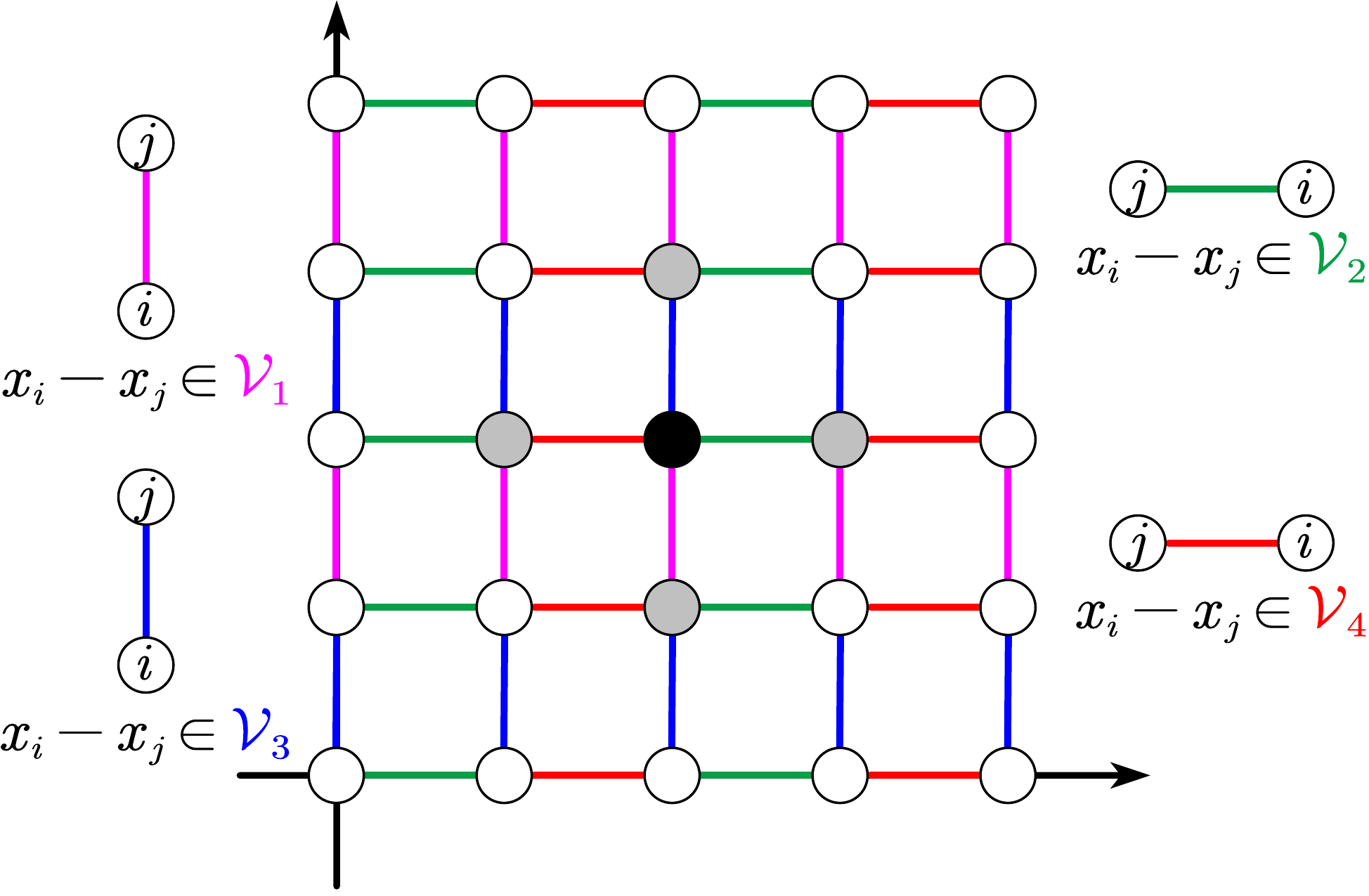}}
\caption{Clique partitioning using the 4-neighbourhood structure. Two edges within the subset $\mathcal{V}_k$ ($k=1,2,3,4$) cannot share a vertex (pixel). Each pixel is involved in four edges (one edge in each $\mathcal{V}_k$).}
\label{fig:illustration_4_connectivity}
\end{figure}

In this work, we consider, as in \cite{muntoni2019nonconvex}, a set of prior distributions based on the expected properties of the discrete gradients of natural images. 
The gradients of a 2D gray scale image $\boldsymbol x$
are defined as the difference between adjacent pixels along the horizontal and vertical directions. To keep the inference process scalable, we consider priors which can be expressed, using the Hammersley-Clifford theorem \cite{clifford1990markov}, as 
\begin{eqnarray}
f_x(\boldsymbol x|\boldsymbol \theta)\propto\prod_{(i,j)\in \mathcal{V}} \phi(x_i-x_j;\boldsymbol \theta),
\label{eq:prior0}
\end{eqnarray}
where $\mathcal{V}$ denotes the set of pairs of pixels that are direct neighbours (with a 4-neighbourhood structure) and $\phi(\cdot;\boldsymbol\theta)$ is a positive function parameterised by $\boldsymbol\theta$. This factorization allows each factor in \eqref{eq:prior0} to only depend on two pixels, which in turn enables efficient EP updates. Moreover, the set of cliques in $\mathcal{V}$ can be partitioned into a four (disjoined) sets of cliques, denoted by $\{\mathcal{V}_k\}_{k=1,2,3,4}$, such that each pixel appears at most once in the list of pixel pairs in $\mathcal{V}_k, \forall k$. Using this partitioning, as illustrated in Fig. \ref{fig:illustration_4_connectivity}, the prior in  \eqref{eq:prior0} is factorized as 
\begin{equation}
    f_x(\boldsymbol x|\boldsymbol\theta) = \frac{1}{C(\boldsymbol\theta)} \prod\limits_{k=1}^4\phi_{{\mathcal V}_k} (\boldsymbol x|\boldsymbol\theta), 
    \label{eq:prior1}
\end{equation}
where $\phi_{{\mathcal V}_k} (\boldsymbol x|\boldsymbol\theta)\propto \prod_{(i,j)\in \mathcal{V}_k} \phi(x_i-x_j;\boldsymbol\theta)$. 

We consider three parametric functions for $\phi(\cdot;\boldsymbol\theta)$, leading to three gradient-based priors. 


\underline{$\ell_1$-TV:} the first type of $\phi(\cdot;\boldsymbol\theta)$ is given by
\begin{equation}
  \phi(x_i-x_j;\boldsymbol\theta)=\exp(-\lambda  \abs{x_i-x_j}),
\label{Eq: L1_TV}  
\end{equation}
where $\boldsymbol\theta := \lambda\geq 0$ is scalar hyperparameter, resulting in $f_x(\boldsymbol x|\boldsymbol\theta)$ in \eqref{eq:prior1} being the classical anisotropic ($\ell_1$-norm) TV prior. A convenient property of this gradient-based prior is that it is log-concave (and thus unimodal), which, when combined with the likelihood \eqref{Eq: likelihood}, makes the posterior distribution of $\boldsymbol x$ log-concave. Note that \eqref{eq:prior0} cannot be used directly to model the isotropic TV-based prior.

In natural images, the gradients can present different (local) distributions in homogeneous regions and at the boundaries of objects. It can be difficult using $\ell_1$-TV prior to recover simultaneously sharp boundaries and large, textured regions. For such tasks, it is preferable to use more aggressive and flexible distributions. In the following we consider Gaussian mixtures, which leads to closed-form EP updates and good performance in practice. 

\underline{MoG2-TV prior:} the second type of
$\phi(\cdot;\boldsymbol\theta)$ is constructed using a mixture of two Gaussian (MoG2) profiles, i.e., 
\begin{equation}
\begin{aligned}
\phi(x_i-x_j;\boldsymbol\theta) =& \omega{\mathcal N}(x_i-x_j;0,s_1^2)\\
&+ (1-\omega){\mathcal N}(x_i-x_j;0,s_2^2),
\end{aligned}
\label{Eq: MoG2_TV_prior}
\end{equation}
where $\boldsymbol\theta:=(\omega,s_1^2,s_2^2)$ includes three hyperparameters $\omega$, $s_1$, and $s_2$. Without loss of generality, we assume $s_1>s_2>0$. The first Gaussian term encodes the distribution of the large image gradients expected at object boundaries, while the second term represents the distribution of image gradients within homogeneous regions (where image gradients are expected to be smaller). The parameter $\omega \in (0,1)$ controls the prior fraction of small/large image gradients. The mixture of more than two Gaussian distributions could also be used, however it would introduce additional hyperparameters whose setting would remain challenging. 

\underline{BG-TV prior:} the last type of $\phi(\cdot;\boldsymbol\theta)$ considered is obtained by letting $s_2^2$ above tend to 0. In that case, $\phi(\cdot;\boldsymbol\theta)$ reduces to a Bernoulli-Gaussian (BG) mixture, whereby image gradients are a priori either exactly zero, or Gaussian distributed, i.e., 
\begin{equation}
\phi(x_i - x_j;\boldsymbol\theta) = \omega{\mathcal N}(x_i - x_j;0,s^2)+(1-\omega) \delta(x_i - x_j), 
\label{Eq: BG_TV}
\end{equation}
where $\boldsymbol\theta:=(\omega,s^2)$. Although this prior seems very informative/restrictive, it only depends on two hyperparameters, which can be easier to tune than those involved in \eqref{Eq: MoG2_TV_prior} and can be appropriate when the scene of interest presents piece-wise constant intensity profiles \cite{muntoni2019nonconvex}. 
\subsection{Exact posterior distribution}


Irrespective of the form of $\phi(\cdot;\boldsymbol\theta)$ in \eqref{Eq: L1_TV} - \eqref{Eq: BG_TV}, using the factorization in \eqref{eq:prior1} and the Bayes rule, the posterior distribution of $\boldsymbol x$ conditioned on $\boldsymbol\theta$ is given by
\begin{equation}
    f(\boldsymbol x|\boldsymbol y,\boldsymbol\theta) = \frac{f_y(\boldsymbol y|\textbf{H}\boldsymbol x)f_x(\boldsymbol x|\boldsymbol\theta)}{\int f_y(\boldsymbol y|\textbf{H}\boldsymbol x)f_x(\boldsymbol x|\boldsymbol\theta) {\rm d}\boldsymbol x}.
\label{Eq: exact_posterior_distribution}    
\end{equation}
Bayesian inference based on $f(\boldsymbol x|\boldsymbol y,\boldsymbol\theta)$ is usually intractable as the denominator in \eqref{Eq: exact_posterior_distribution} is typically intractable. Although sampling from the posterior is possible, in particular for the $\ell_1$-TV prior where the posterior is log-concave \cite{durmus2018efficient, vargas2019accelerating}, it remains challenging in high-dimensional settings, in particular with the multimodal priors induced by \eqref{Eq: MoG2_TV_prior} and \eqref{Eq: BG_TV}.


\subsection{Extended Bayesian model and EP posterior approximation}
Although the proposed EP method can be applied directly to approximate the posterior distribution in \eqref{Eq: exact_posterior_distribution}, we introduce an extended model including a set of auxiliary variables gathered in a set denoted by $\boldsymbol u = \{u_n\}_{n=1}^{2N}$. $\boldsymbol u$ contains all the vertical and horizontal gradients of $\boldsymbol x$, i.e., $ \boldsymbol u = \{x_i-x_j\}_{(i,j) \in \mathcal{V}}$. Using the partitioning described in Section \ref{Subsec: TV_priors}, $\boldsymbol u$ can be partitioned as $\boldsymbol u=\{\boldsymbol u_1,\boldsymbol u_2,\boldsymbol u_3,\boldsymbol u_4\}$, such that $\boldsymbol u_k = \{x_i-x_j\}_{(i,j) \in \mathcal{V}_k}$, $\forall k$. One of the key ingredients of the proposed EP algorithms is the one-to-one mapping between $\boldsymbol u_k$ and the pixel pairs in ${\mathcal V}_k$, i.e., $u_n = x_i - x_j$. 
The introduction of $\boldsymbol u$ (\romannumeral 1) does not change the approximation of the posterior of $\boldsymbol x$, (\romannumeral 2) generalizes the description of the proposed EP methods as the same unifying equations and updates can be used irrespective of $\phi(\cdot;\boldsymbol\theta)$, and (\romannumeral 3) allows to efficiently estimate $\lambda$ (if unknown) when using the $\ell_1$-TV prior (see Section \ref{Sec: EP-EM for hyperparameter estimation of TV-L1 prior}).

In the extended Bayesian model, the prior of $\boldsymbol u$ conditioned on $\boldsymbol x$, is defined as $f_u(\boldsymbol u|\boldsymbol x)=\prod_{k=1}^4 f_u(\boldsymbol u_k|\boldsymbol x)$, where 
\begin{equation}
  f_u(\boldsymbol u_k|\boldsymbol x)=\delta(\boldsymbol u_k - {\mathbf  D}_k\boldsymbol x),  \,\, \forall k \in \{1,2,3,4\}.
  \label{eq:prior_u}
\end{equation}
The $N/2 \times N$ matrix ${\mathbf  D}_k$ allows the computation of the gradients of $\boldsymbol x$ associated with the edges in $\mathcal{V}_k$, and $\delta(\cdot)$ denotes a product of Dirac delta functions, applied element-wise to its (multivariate) input. Eq. \eqref{eq:prior_u} ensures that the elements of $\boldsymbol u_k$  correspond to the gradients of $\boldsymbol x$ in ${\mathcal V}_k$. Combining $f_x(\boldsymbol x|\boldsymbol\theta)$ in \eqref{eq:prior1} and $f_u(\boldsymbol u|\boldsymbol x)$, the resulting extended posterior distribution becomes 
\begin{equation}
    f(\boldsymbol x,\boldsymbol u|\boldsymbol y,\boldsymbol\theta) \propto f_y(\boldsymbol y|{\mathbf H}\boldsymbol x) \prod\limits_{k=1}^4[\phi_{{\mathcal V}_k} (\boldsymbol x|\boldsymbol\theta) f_{u}(\boldsymbol u_k|\boldsymbol x)].
\label{Eq: extended_posterior}    
\end{equation}
We then approximate the extended posterior in \eqref{Eq: extended_posterior} using a Gaussian distribution $Q(\boldsymbol x,\boldsymbol u)$\footnotemark[3] such that
\begin{equation}
Q(\boldsymbol x,\boldsymbol u) \approx f(\boldsymbol x,\boldsymbol u|\boldsymbol  y, \boldsymbol\theta).
\end{equation}\footnotetext[3]{To simplify the notation,  $Q(.)$ is used to represent multivariate Gaussian distributions with respect to (w.r.t.) $\boldsymbol x$, $(\boldsymbol x,\boldsymbol u)$, or $\boldsymbol u$ in the remainder of this paper.}

To ensure the proposed EP algorithm remains tractable, a classical mean-field approximation framework \cite{parisi1988statistical, bishop2006pattern} is used to factorize $Q(\boldsymbol x ,\boldsymbol u)$ such that
\begin{equation}
\label{eq:mean_field_Q}
 Q(\boldsymbol x,\boldsymbol u) = Q(\boldsymbol x)Q(\boldsymbol u),   
\end{equation}
with $Q(\boldsymbol x) = {\mathcal N}(\boldsymbol x;\boldsymbol\mu_x, {\mathbf \Sigma}_x)$ and  $Q(\boldsymbol u) = {\mathcal N}(\boldsymbol u;\boldsymbol \mu_u, {\mathbf \Sigma}_u)$. This approximation basically decouples the approximate distribution of the image $\boldsymbol x$ from that of its gradients (encoded in $\boldsymbol u$). Note that the marginal distribution obtained by integrating $f(\boldsymbol x,\boldsymbol u|\boldsymbol y,\pmb\theta)$ over $\boldsymbol u$ is the original posterior distribution in \eqref{Eq: exact_posterior_distribution}. Thus, using the marginal posterior approximation $\int Q(\boldsymbol x,\boldsymbol u) \textrm{d}\boldsymbol u$ is equivalent to approximating the original posterior in \eqref{Eq: exact_posterior_distribution}. Moreover, to keep Bayesian inference scheme scalable and numerically stable, the covariance matrices ${\mathbf \Sigma}_x$ and ${\mathbf \Sigma}_u$ are enforced to be diagonal.

\section{Proposed EP algorithm with TV priors}
\label{Sec: Proposed EP algorithm with TV priors}
This section proposes new EP algorithm to find the two Gaussian densities $Q(\boldsymbol x)$ and $Q(\boldsymbol u)$.
\subsection{EP factorization}
\begin{figure}[ht!]
\centerline{\includegraphics[width=0.45\textwidth]{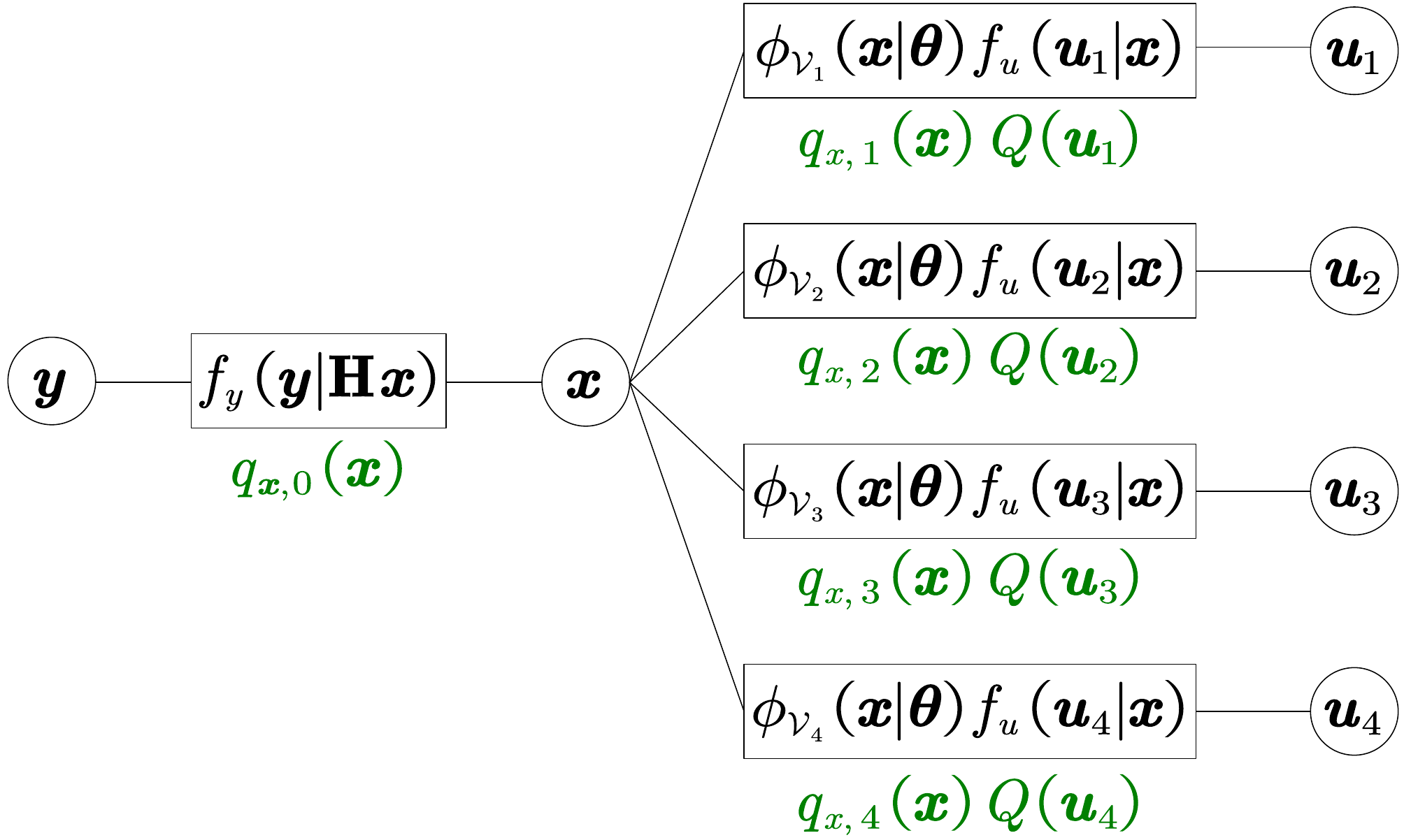}}
\caption{Factor graph used to perform EP approximation for extended exact posterior distribution in \eqref{Eq: extended_posterior}. The rectangle boxes (resp. circles) represent the factor (resp. variables) nodes and EP approximating distribution for each factor node is shown in green.}
\label{fig:factor_graph}
\end{figure}
EP leverages the factorization of the exact posterior in \eqref{Eq: extended_posterior}, which reduces to five factors depicted in Fig. \ref{fig:factor_graph}. More precisely, each exact factor is associated with an approximating factor such that
\begin{eqnarray}
q_{x,0}(\boldsymbol x)  & \approx & K_0 f_y(\boldsymbol y|{\mathbf H}\boldsymbol x),\nonumber\\
q_{x,k}(\boldsymbol x)Q(\boldsymbol u_k)  & \approx & K_k \phi_{{\mathcal V}_k} (\boldsymbol x|\boldsymbol\theta)f_u(\boldsymbol u_k|\boldsymbol x), \forall k \in \{1,\ldots,4\}, \nonumber
\end{eqnarray}
where $\{K_k\}_{k=0}^4$ are constants and the approximating factors are Gaussian densities whose moments are denoted as follows
\begin{equation*}
\begin{aligned}
\begin{cases}
q_{x,k}(\boldsymbol x)= {\mathcal N}(\boldsymbol x;\boldsymbol\mu_{x,k},{\mathbf \Sigma}_{x,k}), \quad\forall k \in \{0,\ldots,4\},\\
Q(\boldsymbol u_k) = {\mathcal N}(\boldsymbol u_k; \boldsymbol\mu_{u,k}, {\mathbf \Sigma}_{u,k}), \quad\forall k \in \{1,\ldots,4\}.
\end{cases}
\end{aligned}
\end{equation*}
The \emph{global} EP marginal approximations $Q(\boldsymbol x)$ and $Q(\boldsymbol u)$ in \eqref{eq:mean_field_Q} can then be obtained by 
\begin{equation}
    Q(\boldsymbol x) \propto q_{x,0}(\boldsymbol x)\prod\limits_{k=1}^4 q_{x,k}(\boldsymbol x), \quad Q(\boldsymbol u) = \prod\limits_{k=1}^4 Q(\boldsymbol u_k).
\end{equation}
To ensure that ${\mathbf \Sigma}_x$ and ${\mathbf \Sigma}_u$ are diagonal, $\{{\mathbf \Sigma}_{x,k}\}_{k=0}^4$ and $\{{\mathbf \Sigma}_{u,k}\}_{k=1}^4$ are forced to be diagonal (and positive-definite) during the EP updates, as will be discussed in subsection \ref{subsec: Update rules for approximating factors}.

\subsection{KL divergence minimization}
EP based on Gaussian approximations can be seen as an sequential message passing algorithm which, at each iteration, updates the means and covariance matrices of the five EP approximating factors in Fig. \ref{fig:factor_graph}. More precisely, each iteration consists of solving sequentially the following KL divergence minimization problems
\begin{subequations}
\begin{equation}
\mathop{\min}\limits_{q_{x,0}(\boldsymbol x)}KL(f_{y}(\boldsymbol y|{\mathbf H}\boldsymbol x)Q^{\backslash 0}(\boldsymbol x)||Q(\boldsymbol x)),
\label{Eq: KL_a} 
\end{equation}
\begin{equation}
\mathop{\min}\limits_{q_{x, k}(\boldsymbol x)Q(\boldsymbol u_k)}KL(\phi_{\mathcal{V}_k}(\boldsymbol x|\boldsymbol\theta)f_u(\boldsymbol u_k|\boldsymbol\theta)Q^{\backslash k}(\boldsymbol x)|| Q(\boldsymbol x)Q(\boldsymbol u_k)),
\label{Eq: KL_b} 
\end{equation}
\label{Eq: KL_minimization}
\end{subequations}
\hspace{-0.9em} for $k \geq 1$, where $\{Q^{\backslash k}(\boldsymbol x)\}_{k=0}^4$ denote the so-called cavity distributions obtained by removing a Gaussian approximating factor from the global posterior approximation $Q(\boldsymbol x)$, i.e., $Q^{\backslash k}(\boldsymbol x) \propto
    Q(\boldsymbol x)/q_{x,k}(\boldsymbol x)$ is the ratio of two Gaussian densities (see \cite[Appendix.~3]{hernandez2015expectation}).
The cavity distributions are also Gaussian distributions, with means and covariance matrices defined by $Q^{\backslash k}(\boldsymbol x) = {\mathcal N}(\boldsymbol x;\boldsymbol\mu_x^{\backslash  k},{\mathbf \Sigma}_x^{\backslash k}), \forall k\geq 0$. Moreover, note that the covariance matrices $\{{\mathbf \Sigma}_x^{\backslash k}\}_k$ are also diagonal by construction.

The first arguments of the KL divergences in \eqref{Eq: KL_minimization} are called tilted distributions, and are formed by the product of an exact factor to be approximated and the corresponding cavity distributions. To ease notations, they are denoted by $P_k(\boldsymbol x,\boldsymbol u), \forall k\geq 0$ (although $P_0(\boldsymbol x,\boldsymbol u)$ does not depend on $\boldsymbol u$). The second arguments of the KL divergences are the global EP approximations or one of its marginals. Since $Q(\boldsymbol x)$ and $Q(\boldsymbol u)$ are Gaussian densities with diagonal covariance matrices, solving the problems in \eqref{Eq: KL_minimization} reduces to matching the marginal moments (means and variances) of $P_k(\boldsymbol x,\boldsymbol u)$ and $Q(\boldsymbol x,\boldsymbol u)$ (see \cite[Chap.~10]{bishop2006pattern}). For instance, to update the factors $q_{x, k}(\boldsymbol x)$ and $Q(\boldsymbol u_k)$ (for $k\geq 1$), one can compute the marginal moments of $P_k(\boldsymbol x,\boldsymbol u)$ w.r.t. $\boldsymbol x$ and $\boldsymbol u_k$. This provides directly the marginal moments of $Q(\boldsymbol  u_k)$ and the updated moments of $Q(\boldsymbol x)$. The updated parameters of $q_{x, k}(\boldsymbol x)$ are then obtained using $q_{x,k}(\boldsymbol x) \propto Q(\boldsymbol x)/Q^{\backslash k}(\boldsymbol x)$. Similarly, updating $q_{x,0}(\boldsymbol x)$ simply requires the computation of the marginal moments (w.r.t. $\boldsymbol x$) of $P_0(\boldsymbol x,\boldsymbol u)$. The next section details the update of the different approximating factors, including the computation of the marginal moments of the different tilted distributions.

\subsection{Updating the approximating factors}
\label{subsec: Update rules for approximating factors}

\textbf{Update of $q_{x,0}(\boldsymbol x)$:} $q_{x,0}(\boldsymbol x)$ is the minimizer of \eqref{Eq: KL_a}, which is used to approximate $f_{y}(\boldsymbol y|{\mathbf H}\boldsymbol x)$. Although the tilted distribution $P_0(\boldsymbol x) = f_y(\boldsymbol y|{\mathbf H}\boldsymbol x)Q^{\backslash 0}(\boldsymbol x)$ is a multivariate Gaussian distribution whose mean and covariance matrix can be obtained in closed-form as follows
\begin{equation}
\begin{aligned}
\begin{cases}
{\rm Cov}_{P_0}(\boldsymbol x) = \left[\frac{1}{\xi} {\mathbf H}^T {\mathbf H}+({\mathbf \Sigma}_x^{\backslash 0})^{-1}\right]^{-1},\\
{\mathbb E}_{P_0}[\boldsymbol x] ={\rm Cov}_P(\boldsymbol x)\left[\frac{1}{\xi} {\mathbf H}^T \boldsymbol y + ({\mathbf \Sigma}_{x}^{\backslash 0})^{-1}\boldsymbol\mu_{x}^{\backslash 0}\right],\\
\end{cases}
\end{aligned}
\label{Eq: tilted_P_1}
\end{equation}
the computation of ${\rm Cov}_{P_0}(\boldsymbol x)$ can be computationally expensive (due to the matrix inversion) depending on the  structure of ${\mathbf H}{\transp} {\mathbf H}$. This costly matrix inversion with complexity ${\mathcal O}(N^3)$ over multiple iterations remains the computational bottleneck of EP algorithms in general \cite{hernandez2015expectation}. Computing ${\mathbb E}_{P_0}[\boldsymbol x]$ in the second line of \eqref{Eq: tilted_P_1} is less challenging as it can be achieved efficiently using conjugate gradient methods \cite{hestenes1952methods} given that left-multiplying by ${\rm Cov}_{P_0}^{-1}(\boldsymbol x)$ is often simple. Since $Q(\boldsymbol x)$ has a diagonal covariance matrix, only the diagonal elements of ${\rm Cov}_{P_0}(\boldsymbol x)$ are actually needed. These marginal variances are gathered in a diagonal matrix denoted ${\rm Var}_{P_0}(\boldsymbol x)$, such that ${\rm diag}({\rm Var}_{P_0}(\boldsymbol x) ) = {\rm diag}({\rm Cov}_{P_0}(\boldsymbol x))$. Given that ${\mathbf \Sigma}_x^{\backslash 0}$ is diagonal, if ${\mathbf H}^T {\mathbf H}$ is diagonal, computing ${\rm Cov}_{P_0}(\boldsymbol x)$ is trivial (inversion of a diagonal matrix) and if ${\mathbf H}^T {\mathbf H}$ is low-rank, the Woodbury matrix identity can be used to compute ${\rm Cov}_{P_0}(\boldsymbol x)$. For more general ${\mathbf H}$, ${\rm diag}({\rm Var}_{P_0}(\boldsymbol x))$ can be approximated via Monte Carlo sampling, e.g., Rao-Blackwellized Monte Carlo (RBMC) method proposed in  \cite{siden2018efficient}. Note that if the noise is not i.i.d., its covariance matrix can be easily integrated in \eqref{Eq: tilted_P_1}, which makes the denoising problem with non i.i.d. noise trivial using our EP methods.

Once ${\rm Var}_{P_0}(\boldsymbol x)$ and ${\mathbb E}_{P_0}[\boldsymbol x]$ are computed, the parameters of $q_{x,0}(\boldsymbol x)$ can be updated via
\begin{equation}
\begin{aligned}
&{\mathbf \Sigma}_{x,0} = \left[\left({\rm Var}_{P_0}(\boldsymbol x)\right)^{-1} - ({\mathbf \Sigma}_x^{\backslash 0})^{-1}\right]^{-1},\\
&\boldsymbol\mu_{x,0} = {\mathbf \Sigma}_{x,0} \left[\left({\mathbf \Sigma}_{x,0}^{-1}+({\mathbf \Sigma}_x^{\backslash 0})^{-1}\right) \boldsymbol{\mathbb E}_{P_0}[\boldsymbol x] - ({\mathbf \Sigma}_x^{\backslash 0})^{-1}\boldsymbol\mu_{x,0}^{\backslash 0}\right].\\
\end{aligned}
\label{Eq: update_q0_x}  
\end{equation}

\textbf{Update of $q_{x,k}(\boldsymbol x)$ and $Q(\boldsymbol u_k)$:} Update of these approximating factors requires the computation of the marginal moments of 
tilted distribution $ P_k(\boldsymbol x, \boldsymbol u_k) = \phi_{\mathcal{V}_k}(\boldsymbol x|\boldsymbol\theta)f_u(\boldsymbol u_k|\boldsymbol\theta)Q^{\backslash k}(\boldsymbol x)$ in \eqref{Eq: KL_b}, $\forall k \geq 1$. It can be shown that marginalizing over $\boldsymbol u_k$ leads to 
\begin{equation}
\begin{aligned}
P_k(\boldsymbol x) &= \int P_k(\boldsymbol x,\boldsymbol u_k) {\rm d}\boldsymbol u_k =  \prod_{(i,j) \in {\mathcal V}_k} P_k(x_i,x_j)  \\
&=\prod_{(i,j) \in {\mathcal V}_k}  \phi (x_i -x_j;\boldsymbol\theta)Q_x^{\backslash k}(x_i)Q_x^{\backslash k}(x_j),
\label{Eq: marginal_tilted_x}
\end{aligned}
\end{equation}
where $Q_{x}^{\backslash k}(x_i)$ and $Q^{\backslash k}_{x}(x_j)$, whose mean and variance are $(m_i,c_i)$ and  $(m_j,c_j)$, denote the marginals of $x_i$ and $x_j$ associated with $Q_x^{\backslash k}(\boldsymbol x)$. Moreover, we define $Q_{x}^{\backslash k}(x_i,x_j)=Q_{x}^{\backslash k}(x_i)Q_{x}^{\backslash k}(x_j)$, where $x_i$ and $x_j$ are pixel pairs appearing in a single pair/edge in ${\mathcal V}_k$. Thus, $P_k(\boldsymbol x)$ is the product of $N/2$ two-dimensional unnormalized densities $P_k(x_i,x_j)$ associated with the $N/2$ pixel pairs in ${\mathcal V}_k$.

Similarly, it can be shown that marginalizing over $\boldsymbol x$ yields
\begin{equation}
P_k(\boldsymbol u_k) = \int P_k(\boldsymbol x,\boldsymbol u_k) {\rm d}\boldsymbol x  = \prod_{n \in {\mathcal V}_k} P_k(u_n).
\label{Eq: marginal_tilted_u}
\end{equation}
The update of $Q(\boldsymbol u_k)$ requires computing the moments of $P_k(\boldsymbol u_k)$, where 
\begin{equation}
P_k(u_n) \propto {\mathcal N}(u_n;m_i-m_j, c_i+c_j) \phi(u_n;\boldsymbol\theta).
\label{Eq: marginal_tilted_un}
\end{equation}
For the three priors considered, $P_k(u_n)$ reduces to simple mixtures of two distributions involving (truncated) Gaussian distributions. Thus, its mean $\bar{m}_n$ and variance $\bar{S}_n$ can be computed easily. Prior-dependent expressions are omitted here for brevity.
If more exotic functions were used for $\phi(\cdot;\boldsymbol\theta)$, the moments of $P_k(u_n)$ could still be computed, e.g., via numerical integration. 

As for the update of $q_{x,k}(\boldsymbol x)$, it requires computing the moments of $P_k(\boldsymbol x)$ in \eqref{Eq: marginal_tilted_x}. Using the projection scheme in \cite{gehre2014expectation}, the marginal moments of $P_k(x_i,x_j)$ can be obtained via 
\begin{equation*}
\begin{aligned}
{\mathbb E}_{P_k}[(x_i, x_j)] &= (m_i, m_j) +  \frac{\bar{m}_n-\left( m_i-m_j \right)}{c_i+c_j} (c_i, -c_j),\\
{\rm Var}_{P_k}(x_i, x_j) &= (c_i, c_j) + \frac{\xbar {S}_n - (c_i +c_j)}{(c_i + c_j)^2}(c_i^2, c_j^2).
\end{aligned}
\label{Eq: tilted_mean_var_proj}
\end{equation*}
\noindent It can be seen from \eqref{Eq: marginal_tilted_u} that the moments of the $N/2$ marginals w.r.t. $\boldsymbol u_k$ can all be computed independently (i.e., in parallel),  after which all the edges/pairs $x_i, x_j$ in \eqref{Eq: marginal_tilted_x} can be processed independently as well. Once the marginal moments of ${\rm Var}_{P_k}(\boldsymbol x)$ and ${\mathbb E}_{P_k}[\boldsymbol x]$ for $x_i, x_j \in {\mathcal V}_k$ are computed, the parameters of $q_{x,k}(\boldsymbol x)$ can be obtained via
\begin{equation}
\begin{aligned}
&{\mathbf \Sigma}_{x,k} = \left[{\rm Var}_{P_k}(\boldsymbol x)^{-1} - ({\mathbf \Sigma}_x^{\backslash k})^{-1}\right]^{-1},\\
&\boldsymbol\mu_{x,k} = {\mathbf \Sigma}_{x,k} \left[\left({\mathbf \Sigma}_{x,k}^{-1}+({\mathbf \Sigma}_x^{\backslash k})^{-1}\right) \boldsymbol{\mathbb E}_{P_k}[\boldsymbol x] - ({\mathbf \Sigma}_x^{\backslash k})^{-1}\boldsymbol\mu_{x,k}^{\backslash k}\right].\\
\end{aligned}
\label{Eq: update_qk_x}  
\end{equation}
Therefore, the update of $Q(\boldsymbol u_k)$  and $q_{x,k}(\boldsymbol x)$ is very efficient in practice.

As mentioned earlier, EP iterates by updating sequentially $q_{x,0}(\boldsymbol x)$ and $\{q_{x,k}(\boldsymbol x),Q(\boldsymbol u_k)\}_{k=1}^4$ until convergence. The final EP approximation $Q(\boldsymbol x)$ are obtained by
\begin{equation}
\begin{aligned}
&{\mathbf \Sigma}_x = ({\mathbf \Sigma}_{x,0}^{-1} + \sum\limits_{k=1}^4 {\mathbf \Sigma}_{x,k}^{-1})^{-1},\\
&\boldsymbol\mu_x = {\mathbf \Sigma}_x({\mathbf \Sigma}_{x,0}^{-1}\boldsymbol\mu_{x,0} + \sum\limits_{k=1}^4 {\mathbf \Sigma}_{x,k}^{-1}\boldsymbol\mu_{x,k}).\\
\end{aligned}
\label{Eq: Q_x_final}
\end{equation}

The first lines of \eqref{Eq: update_q0_x} and \eqref{Eq: update_qk_x} do not ensure that the updated covariance matrices ${\mathbf \Sigma}_{x,0}$, ${\mathbf \Sigma}_{x,k}$ are strictly positive definite. If negative variances are obtained, these values are usually replaced by large positive values (e.g., $10^8$) before computing $\boldsymbol\mu_{x,0}$ and $\boldsymbol\mu_{x,k}$ \cite{hernandez2015expectation,gelman2014expectation}.
Algorithm \ref{Algo: EP_with_TV_priors} summarizes the pseudo-code for the proposed algorithm. In general, there is no guarantee for the convergence of EP algorithms. A standard damping strategy  \cite{gelman2014expectation, hernandez2015expectation} is used here to prevent potential oscillations between successive iterations. In practice, if $\boldsymbol\theta$ was set reasonably to reflect the expected distribution (scale) of the gradients of $\boldsymbol x$, we did not experienced convergence issues. Such issues may arise if $\boldsymbol\theta$ is set such that the resulting posterior distribution $f(\boldsymbol x|\boldsymbol y,\boldsymbol\theta)$ is extremely ill-conditioned.

Thanks to the auxiliary variables and the priors considered, all EP updates admit closed-form solutions and enable efficient parallel computation.
Moreover, the mean and marginal variance of auxiliary variable $\boldsymbol u$ can be further used to estimate the hyperparameter $\lambda$ of $\ell_1$-TV prior under an EP-EM framework, as will be discussed in next section.  

\begin{algorithm}[t]
    \SetKwFunction{isOddNumber}{isOddNumber}
    \SetKwInOut{KwIn}{Input}
    \SetKwInOut{KwOut}{Output}

    \KwIn{observation $\boldsymbol y$, degradation matrix ${\mathbf H}$, noise variance $\xi$, hyperparameter $\boldsymbol\theta$}

    \KwOut{$Q(\boldsymbol x) \propto {\mathcal N}(\boldsymbol x;\boldsymbol\mu_x,{\mathbf \Sigma}_x)$}

    $initialization:$ $\boldsymbol\mu_{x,0} = \boldsymbol y$, ${\mathbf \Sigma}_{x,0} = \xi {\mathbf I}_N$, $\{\boldsymbol\mu_{x,k}\}_{k=1,2,3,4} = \boldsymbol 0$,  $\{{\mathbf \Sigma}_{x,k}\}_{k=1,2,3,4} =10^8{\mathbf I}_N$

    \For{t:=1 to StopRule}{
        Compute ${\mathbb E}_{P_0}[\boldsymbol x]$ and ${\rm Var}_{P_0}(\boldsymbol x)$  as in \eqref{Eq: tilted_P_1}. \\
        Update ${\mathbf \Sigma}_{x,0}$ and ${\boldsymbol\mu}_{x,0}$  using \eqref{Eq: update_q0_x}.\\
         \For{$k=1,2,3,4$ (random order)}{
        Compute ${\bar m}_n$, ${\xbar S}_n$ for $(i,j)\in {\mathcal V}_k$ in parallel.\\
        Update ${\mathbf \Sigma}_{x,k}$ and $\boldsymbol\mu_{x,k}$ using \eqref{Eq: update_qk_x}. \\
    }}
    Compute ${\mathbf \Sigma}_x$ and $\boldsymbol\mu_x$ as in \eqref{Eq: Q_x_final}.
    \caption{Proposed EP algorithms ($\boldsymbol\theta$ is known)}
\label{Algo: EP_with_TV_priors}    
\end{algorithm}

\section{Unsupervised EP algorithm with $\ell_1$-TV prior}
\label{Sec: EP-EM for hyperparameter estimation of TV-L1 prior}
In this section, we propose to embed the proposed EP algorithm using the $\ell_1$-TV prior within a larger inference problem where the hyperparameter $\lambda$ is unknown and needs to be adjusted for the image of interest. We consider only this prior as it has been shown in  \cite{pereyra2015maximum} that the normalizing constant $C(\lambda)$ in this case has a closed-form expression, leading to 
\begin{equation}
    f(\boldsymbol x|\lambda) = \dfrac{1}{D\lambda^{-N}}\exp\left[-\lambda \sum_{(i,j)\in \mathcal{V}}\abs{x_i-x_j}\right],
\end{equation}
where $D$ is a constant independent of $\lambda$. Although this prior is not proper, the posterior distribution often is in practice. We do not consider MoG2-TV and  BG-TV priors in this section as the associated constants $C(\boldsymbol\theta)$ in those cases are intractable.

It has been shown in \cite{vidal2020maximum} that when the prior is parametrized by a reduced number of hyperparameters (only a scalar-valued $\lambda$ here), estimating these hyperparameters by maximum marginal likelihood estimation (MMLE) or marginal maximum a posteriori (MMAP) estimation tends to lead to better image estimates than  estimating $\boldsymbol x$ and the hyperparameters by joint MMSE estimation. Here, we present a MMLE approach which aims at maximizing  
\begin{equation}
f(\boldsymbol y|\lambda) = \int f_y(\boldsymbol y|{\mathbf H}\boldsymbol x)f_x(\boldsymbol x|\lambda) {\rm d}\boldsymbol x, 
\end{equation}
i.e., the normalizing constant of the exact posterior in \eqref{Eq: exact_posterior_distribution}. However, a similar approach could be used to maximize $f(\lambda|\boldsymbol y)$, in particular if $\lambda$ is assigned a conjugate prior \cite{pereyra2015maximum}.

Maximizing $f(\boldsymbol y|\lambda)$ can be achieved using an iterative procedure based on EM, where the standard update rule at iteration $(t)$ is 
\begin{equation}
    \lambda^{(t)} = \underset{\lambda}{\text{argmax}}~ {\mathbb E}_{f(\boldsymbol x|\boldsymbol y,\lambda^{(t-1)})} [\log f(\boldsymbol x|\boldsymbol y,\lambda)],
\label{Eq: E_step_exact_posterior}    
\end{equation}
yielding the cost function $F^{(t)}(\lambda)$ to be maximized as
\begin{equation}
\hspace{-0.22cm}F^{(t)}(\lambda) = N\log \lambda -\lambda  {\mathbb E}_{f(\boldsymbol x|\boldsymbol y,\lambda^{(t-1)})} \left[\sum_{(i,j)\in \mathcal{V}}\abs{x_i-x_j}\right].
\end{equation}
By zeroing its derivative w.r.t. $\lambda$, the estimate of $\lambda$ at iteration $(t)$ is obtained by $\lambda^{(t)}=N/a_0$, with
\begin{equation}
a_0={\mathbb E}_{f(\boldsymbol x|\boldsymbol y,\lambda^{(t-1)})} \left[\sum_{(i,j)\in \mathcal{V}}\abs{x_i-x_j}\right].
\label{Eq: a_0} 
\end{equation}

\begin{figure}[t]
\centerline{\includegraphics[width=0.47\textwidth]{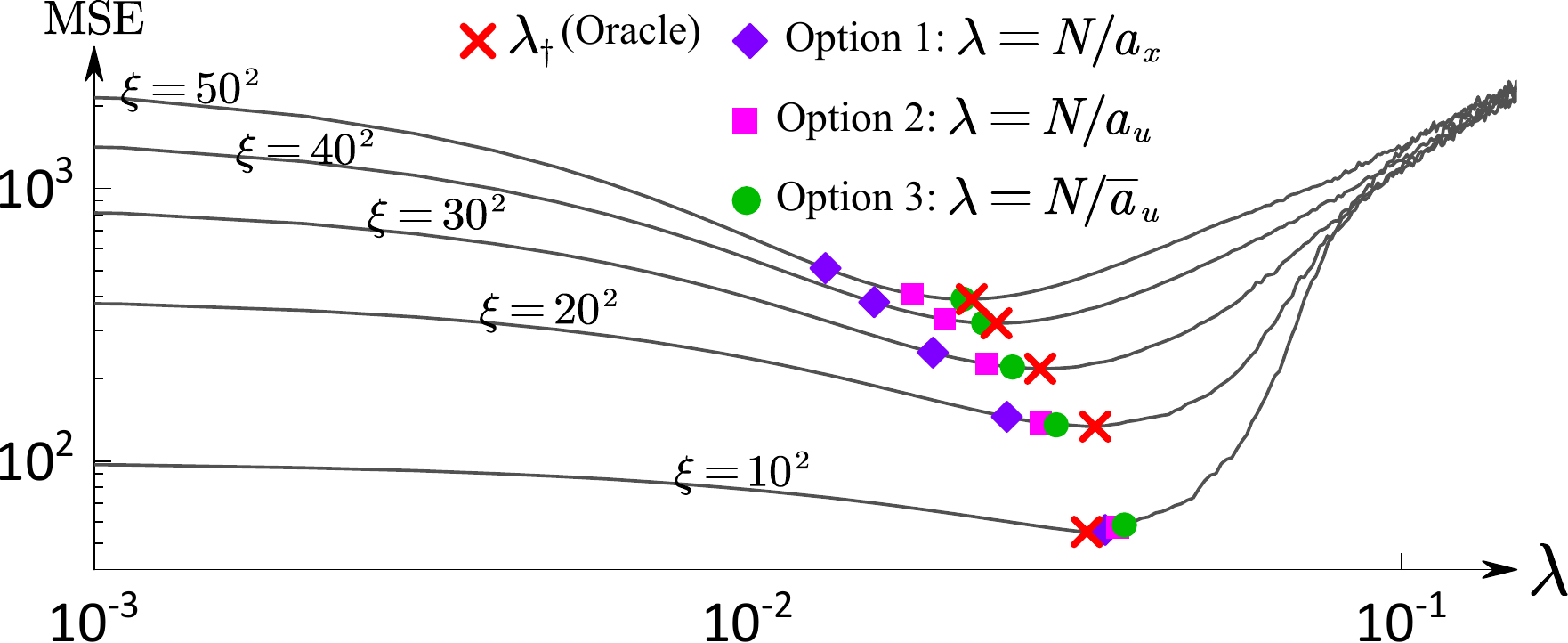}}
\caption{Impact of the strategy used to approximate $a_0$ for a denoising problem ($256\times$256-pixel \textit{Cameraman} image). The MSE (see definition in Section \ref{Sec: Experimental Results}) is presented as a function of $\lambda$ and $\xi$. $\lambda_{\dag}({\rm Oracle})$ is the oracle value of $\lambda$ that minimizes the MSE and the other markers represent the final EP-EM estimates for Options 1-3.}
\label{fig: compare_lam_x_vs_lam_u}
\end{figure}

When exact computation of $a_0$ using
expectation w.r.t. the exact posterior $f(\boldsymbol x|\boldsymbol y,\lambda^{(t-1)})$ is intractable, classical options turn to stochastic approximations (such as stochastic EM variants) or variational approximation (such as variational EM) \cite{celeux2003procedures}. The latter option is adopted here, and the use of the EP posterior approximation leads to an EP-PM algorithm. Three options to approximate $a_0$ in \eqref{Eq: a_0} are considered by replacing directly the expectation w.r.t. $f(\boldsymbol x|\boldsymbol y,\lambda^{(t-1)})$ to another approximate distribution as follows:
\begin{align}
\hspace{-1.3cm}\bullet \,  &{\textrm {Option 1:}} \, a_x={\mathbb E}_{Q(\boldsymbol x|\lambda^{(t-1)})} \left[\sum_{(i,j)\in \mathcal{V}}\abs{x_i-x_j}\right],
\label{eq:ax}\\
\bullet \,  &{\textrm {Option 2:} }\,a_u={\mathbb E}_{Q(\boldsymbol u|\lambda^{(t-1)})} \left[\norm{\boldsymbol u}_{1}\right], \label{eq:au}\\
\bullet \,  &{\textrm  {Option 3:}}\,{\bar a}_u = \sum\limits_{k=1}^4{\mathbb E}_{P_k(\boldsymbol u_k|\lambda^{(t-1)})}\left[\norm{\boldsymbol u_k}_1\right].
\label{Eq: 3rd_alternative_a0}  
\end{align}
All the expectations in \eqref{eq:ax}-\eqref{Eq: 3rd_alternative_a0} can be computed analytically. In practice, \eqref{Eq: 3rd_alternative_a0} performs better for hyperparameter estimation and in turn for unsupervised image restoration than  \eqref{eq:ax} and \eqref{eq:au}, as illustrated via the image denoising examples in Fig. \ref{fig: compare_lam_x_vs_lam_u}. This is due to $Q(\boldsymbol x|\lambda^{(t-1)})$ in \eqref{eq:ax}, which has a diagonal covariance matrix that fails to capture the correlation between adjacent pixels. Using $\sum_{(i,j)\in \mathcal{V}}\abs{x_i-x_j} = \norm{{\boldsymbol u}}_{1}$, \eqref{eq:au} is a better alternative since $Q(\boldsymbol u|\lambda^{(t-1)})$ better captures the posterior variances of the image gradients. Eq. \eqref{Eq: 3rd_alternative_a0} computes expectations w.r.t. tilted distributions, which are expected to be closer to the exact marginal distributions of the gradients than their Gaussian approximations.
Thus \eqref{Eq: 3rd_alternative_a0} is adopted in our EP-EM algorithm whose pseudo-code  is presented in  Algorithm \ref{Algo: unsupervised_EP_EM}. The algorithm can be stopped after a fixed number of iterations or when $\lambda$ stabilises (within 10 to 20 iterations in our experiments, depending on the problem considered). In practice, we observed that a faster version of EP-EM can provide similar results at a lower cost. Instead of running Algorithm \ref{Algo: EP_with_TV_priors} until convergence (line 2 of Algorithm \ref{Algo: unsupervised_EP_EM}), it can be run only for a few iterations (in Section \ref{Sec: Experimental Results}, we used a single iteration). This generally allows us to reduce the overall number of EP iterations and seems more robust if the scale of $\lambda^{(0)}$ is set inappropriately.

\begin{algorithm}[t]
    \SetKwFunction{isOddNumber}{isOddNumber}
    \SetKwInOut{KwIn}{Input}
    \SetKwInOut{KwOut}{Output}

    \KwIn{observation $\boldsymbol y$, degradation matrix ${\mathbf H}$, $\xi$, $\lambda^{(0)}$}
    \KwOut{$Q(\boldsymbol x) \propto {\mathcal N}(\boldsymbol x;\boldsymbol\mu_x,{\mathbf \Sigma}_x)$, $\lambda$ }

    \For{t:=1 to StopRule}{
    Obtain $Q^{(t)}(\boldsymbol x|\boldsymbol y,\lambda^{(t-1)})$ and $\{P^{(t)}_k(\boldsymbol u_k| \lambda^{(t-1)})\}_{k=1}^4$ using Algorithm \ref{Algo: EP_with_TV_priors}\\
        Compute $\bar{a}_u = \sum\limits_{k=1}^4 {\mathbb E}_{P^{(t)}_k(\boldsymbol u_k| \lambda^{(t-1)})}[\norm{\boldsymbol u_{k}}_1]$\\
        Set $\lambda^{(t)} = N/{\bar{a}_u}$.
    }
    \caption{Proposed EP-EM algorithm-unknown $\lambda$}
\label{Algo: unsupervised_EP_EM}    
\end{algorithm}

\section{Experimental Results}
\label{Sec: Experimental Results}

This section evaluates the performance of the proposed algorithms for image denoising, non-blind deconvolution, and CS reconstruction.  
The EP posterior mean $\boldsymbol\mu_x$ is used as image estimate $\hat {\boldsymbol x}$, and the marginal variances in ${\rm diag}({\mathbf \Sigma}_x)$ quantify the pixel-wise posterior uncertainty. The quality of the restored images is assessed using the mean squared error (MSE) and peak signal-to-noise ratio (PSNR) computed between any estimated image $\hat{\boldsymbol x}$ and the corresponding ground truth $\boldsymbol x$, i.e.,
\begin{equation*}
    {\rm MSE} = \frac{1}{N}\norm{\boldsymbol x - \hat{\boldsymbol x}}_2^2, \quad {\rm PSNR} = 10\times \log_{10}\left(\frac{{\rm max}_{\boldsymbol x}^2}{\rm MSE}\right).
\end{equation*}
We will first consider the three gradient-based priors discussed in Section \ref{Sec: Proposed EP algorithm with TV priors} for image denoising to show the effectiveness of the priors. In this context, the hyperparameters $\boldsymbol\theta$ are set using grid search to minimize the MSE, i.e.,
\begin{equation*}
    \boldsymbol\theta_{\dag}({\rm Oracle}) = \underset{\boldsymbol\theta}{\text{argmin}}~ \left\{\norm{\boldsymbol x -  \hat{\boldsymbol x}(\boldsymbol y, \boldsymbol\theta)}_2^{2}\right\},
\end{equation*}
where $\hat{\boldsymbol x}(\boldsymbol y, \boldsymbol\theta)$ is the EP-based image estimate. For completeness, we will also consider (in Fig. \ref{fig: result_deconv_mse_lam}) a second oracle estimator of $\boldsymbol\theta$, where $\hat{\boldsymbol x}(\boldsymbol y, \boldsymbol\theta)$ will be the MAP estimator of $\boldsymbol x$ based on the exact posterior distribution. 
\begin{figure}[b]
\centerline{\includegraphics[width=1\textwidth]{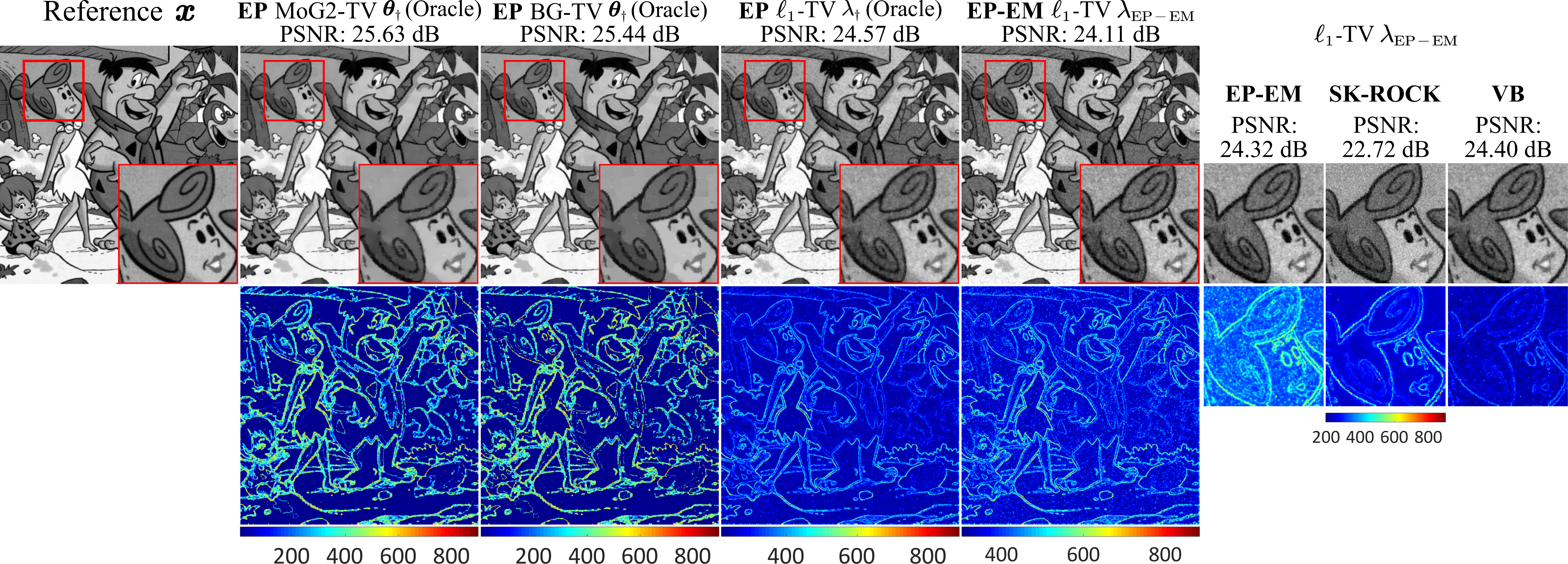}}
\caption{Denoising results of \textit{Flinstones} image ($\xi=30^2$). 2nd-5th columns: results obtained by EP with MoG2-TV, BG-TV, $\ell_1$-TV priors, and EP-EM with $\ell_1$-TV prior. Top row: denoised images. Bottom row: UQ maps. 6th-8th columns: a cropped results by EP-EM, SK-ROCK, and VB with $\ell_1$-TV prior using $\lambda_{\rm EP-EM}$.}
\label{fig:result_denoising_UQ}
\end{figure}

The damping parameter of EP is set to be 0.9 and it has been observed in our experiments that EP converges within 20 iterations. For EP-EM, unless otherwise stated, the results presented are obtained with the fast alternative with 20 EM iterations and a single EP loop per EP-EM iteration. The pixel-wise uncertainty is then represented as a 2D image/map. The main methods used for comparison using the $\ell_1$-TV prior are VB \cite{ruiz2015variational} and SK-ROCK \cite{vargas2019accelerating}, which can be used get approximate MMSE estimates and posterior variances, assuming $\lambda$ is known. Our EP-EM algorithm is also compared to the empirical Bayes (EB) method from \cite{vidal2020maximum}. The SK-ROCK and EB results are considered as reference and for these MCMC algorithms, the burn-in period and chain length (after burn-in) are set to $(2000;10^5)$ samples and $(300;1500)$ samples, respectively. Short chains are used for EB as it is simply used for hyperparameter estimation. While exact MAP estimation using the multimodal priors is not easily achievable, for completeness we will use SALSA \cite{afonso2010fast} for MAP estimation with the $\ell_1$-TV prior.

\subsection{Image denoising}

We first discuss the use of the three priors within EP (Algorithm \ref{Algo: EP_with_TV_priors}) for image denoising problems, i.e., with $\textbf{H}={\mathbf I}_N$. We also include EP-EM (Algorithm \ref{Algo: unsupervised_EP_EM}) to assess its hyperparameter estimation performance.  Experiments are conducted using three widely used grayscale images scaled in $[0, 255]$ ($I_1$: \textit{Cameraman}  256$\times$256 pixels, $I_2$: \textit{Flinstones} 512$\times$512 pixels and $I_3$: \textit{Mandrill}  $512\times$512 pixels). 


The PSNRs obtained after denoising with hyperparameters set to be $\boldsymbol\theta_{\dag}(\rm Oracle)$ are reported in 3rd-5th columns of Table \ref{Table: result_denoising}, for three noise variances. Note that for the three prior models and images, $\boldsymbol\theta_{\dag}(\rm Oracle)$ changes depending on the noise variance, which highlights the need for automatic hyperparameter setting. The two non-convex priors generally present comparable performance (MoG2-TV is marginally better in the high PSNR regime) and provide better PSNRs than the $\ell_1$-TV prior. However, as mentioned before, automatic tuning of the non-convex prior hyperparameters is more challenging than when using the $\ell_1$-TV prior. The PSNRs obtained using EP-EM with the $\ell_1$-TV prior are depicted in the last column of Table \ref{Table: result_denoising}. Although slightly worse compared to $\boldsymbol\theta_{\dag}(\rm Oracle)$, the results in the two last columns are comparable, illustrating the benefit of the proposed EP-EM strategy.


\begin{table}[t]
\centering
\resizebox{0.5\textwidth}{!}{
\begin{NiceTabular}{ccccccc}
\toprule
\multirow{3}{*}{$\xi$} & \multirow{3}{*}{image} &
\multicolumn{3}{c}{$\boldsymbol\theta_{\dag}(\rm Oracle)$ } &  \multicolumn{1}{c}{ $\lambda_{\rm {EP-EM}}$}  \\
\cmidrule(r){3-5}\cmidrule(l){6-6}
& & EP MoG2-TV     &EP BG-TV  & \multicolumn{1}{c}{EP $\ell_1$-TV}   &   EP-EM $\ell_1$-TV\\
& &  $\boldsymbol\theta = (\omega,s_1^2,s_2^2)$& $\boldsymbol\theta= (\omega, s^2)$ & ($\theta = \lambda$)  & ($\theta = \lambda$) \\
\cmidrule{1-6}
\multirow{6}{*}{$10^2$}&\multirow{2}{*}{$I_1$}    & \textbf{33.09} & \underline{32.81}&31.05 &30.77    \\
& & (0.20, 11, 3400) & (0.85, 2800)&(0.0320)& (0.0377) \\
& \multirow{2}{*}{$I_2$}  &\textbf{31.58} & \underline{31.39}& 30.62 &  30.43   \\
& & (0.25, 11, 1700) & (0.80, 1300)&(0.0360) & (0.0278) \\
& \multirow{2}{*}{$I_3$} &\textbf{29.07}    & \underline{29.02}&  28.90 &  28.85  \\
& & (0.25, 71, 3300) & (0.85, 2100 )&(0.0310)&  (0.0263) \\
\cmidrule{2-6}
\multirow{6}{*}{$30^2$}& \multirow{2}{*}{$I_1$}  & \textbf{27.09}  &\underline{26.94}& 25.55&   25.27  \\
& & (0.25, 11, 4000)&(0.75, 5100)&(0.0330) &  (0.0254)\\
& \multirow{2}{*}{$I_2$}   & \textbf{25.63}& \underline{25.44}&24.57 & 24.11  \\
& & (0.25, 11, 4000) & (0.75, 4000)&(0.0270)&  (0.0195) \\
& \multirow{2}{*}{$I_3$}   & \underline{22.91}& \textbf{22.93} & 22.66& 22.66   \\
& & (0.20, 1, 4000) & (0.80, 4700)&(0.0240)&  (0.0233) \\
\cmidrule{2-6}
\multirow{6}{*}{$50^2$}& \multirow{2}{*}{$I_1$}     & \underline{24.24}  &\textbf{24.62}  & 23.05&  22.87  \\
& & (0.25, 1, 4000)&(0.75, 8000) &(0.0260)   &(0.0213)\\
& \multirow{2}{*}{$I_2$}  & \underline{22.35}& \textbf{22.75}&  21.94 & 21.65\\
& & (0.25, 11, 4000) & (0.75, 8200)&(0.0210)&  (0.0162)\\
& \multirow{2}{*}{$I_3$} &\underline{20.86}  & \textbf{20.88}& 20.77 &  20.75\\
& & (0.20, 1, 4000) & (0.80, 4900)&(0.0230)&  (0.0213)&\\
\bottomrule
\end{NiceTabular}}
\caption{Image denoising: PSNR (dB) after denoising using EP. The highest (resp. second highest) PSNR values in each row are bold (resp. underlined). The values in brackets are the values of the corresponding hyperparameter(s).}
\label{Table: result_denoising}
\end{table}




Fig. \ref{fig:result_denoising_UQ} shows examples of denoising results for the \textit{Flinstones} image. The denoised images obtained by EP with the MoG2-TV and BG-TV priors present better visual quality and lower uncertainties in homogeneous regions than using the $\ell_1$-TV prior. The UQ maps obtained with the non-convex priors also present higher contrast between homogeneous regions and object boundaries, mainly because these priors penalise more strongly intermediate gradients. In order to verify the accuracy of UQ maps obtained with the $\ell_1$-TV prior, EP-EM is compared with SK-ROCK and VB for a cropped \textit{Flinstones} image (only a $128\times 128$ pixels portion to keep the processing time of SK-ROCK relatively short). For fair comparisons, SK-ROCK and VB are run with the final hyperparameter estimated via EP-EM, and denoted $\lambda_{\textrm{EP-EM}}$. The images denoised by EP-EM and VB are visually similar and present a slightly higher PSNR than the approximate MMSE estimate obtained via SK-ROCK. Although SK-ROCK can include a bias such that the image reported is not the true MMSE estimate, it is it more likely that this difference is mostly due to the biases of EP and VB, which are beneficial here and yield better PSNRs. When comparing the UQ maps obtained by EP, VB and SK-ROCK (using SK-ROCK as reference), we observe that EP tends to overestimate the uncertainties in homogeneous regions  and extend the high-uncertainty regions at object boundaries. VB tends to underestimate the marginal variances, although they remain close to the SK-ROCK results.

These denoising results illustrate that the MoG2-TV and BG-TV priors can provide better image estimates (when the hyperparameters are correctly set) and we expect similar trends for deconvolution and CS problems. However, hyperparameter tuning by grid search can be computationally intensive and automatic tuning is not straightforward for the two priors. For these reasons, in the remaining experiments, we focus on the $\ell_1$-TV prior, for which comparisons with existing MCMC and VB methods are easier. 

\subsection{Non-blind image deconvolution} 
This subsection illustrates the performance of EP-EM algorithm with $\ell_1$-TV prior for image deconvolution. It is also used to assess its ability to estimate the regularization parameter. Sub-images ($128 \times 128$, $164 \times 164$, $165 \times 165$ pixels) of the three test images $I_1$-$I_3$ are used as reference to keep the computational cost of the competing methods relatively low. The matrix ${\mathbf H}$ corresponds to a $9\times 9$ pixels uniform blurring kernel. The observed images are generated with blurred signal-to-noise-ratio (BSNR) of 15 dB, 25 dB and 35 dB. The alternative hyperparameter estimation methods include EB \cite{vidal2020maximum} and the hierarchical Bayes (HB) method from \cite{pereyra2015maximum}, whose original implementations have been modified to include the $\ell_1$-TV prior. We also consider SUGAR \cite{deledalle2014stein} and the Morozov's discrepancy principle (DP) method \cite{morozov2012methods}, as in \cite{vidal2020maximum}.

\begin{figure}[t]
\centerline{\includegraphics[width=0.7\textwidth]{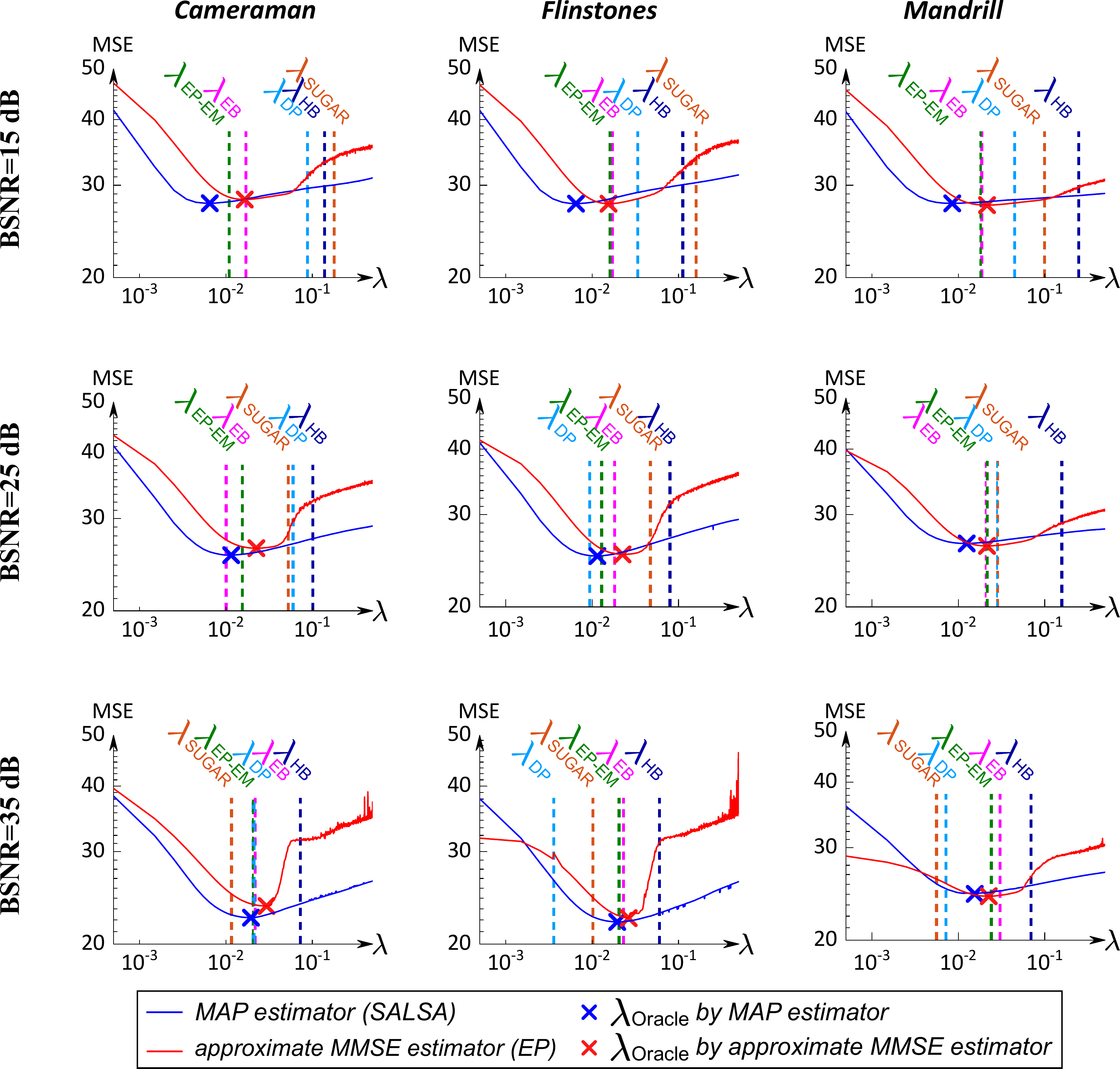}}
\caption{MSE for different $\lambda$ of $\ell_1$-TV prior estimated by different methods. The values on x-y axis are presented on a $\log_{10}$ scale.}
\label{fig: result_deconv_mse_lam}
\end{figure}

Fig. \ref{fig: result_deconv_mse_lam} first presents image MSEs obtained via MAP estimation (blue curves) and EP (approximate MMSE estimation, red curves), as a function of $\lambda$.  We did not include the MSE curves associated with {\textquoteleft  exact\textquoteright}   MMSE estimation, which could be approached via SK-ROCK as it would require running too many Markov chains (the SK-ROCK parameters would also need to be tuned over the range of $\lambda$ considered). For a given value of $\lambda$, the MSE of the MAP estimator is generally lower than that of EP and it is interesting to observe that each of the blue and red curves does not reach its minimum at the same $\lambda$. EB and EP-EM aim at maximizing the same marginal likelihood and $\lambda_{\rm EP-EM}$ and $\lambda_{\rm EB}$ are thus expected to be close. This is confirmed in Fig. \ref{fig: result_deconv_mse_lam} where they are generally close, although $\lambda_{\rm EP-EM}$ can be smaller than $\lambda_{\rm EB}$. Moreover, $\lambda_{\rm EP-EM}$ and $\lambda_{\rm EB}$ are also close to the Oracle values that minimize the MSE using MAP or approximate MMSE estimation. The main benefit of EP-EM over EB is the computational cost since EP-EM does not require high-dimensional Monte Carlo sampling. This makes EP-EM particularly attractive for fast hyperparameter setting, e.g., for subsequent use with MAP-based algorithm.



\begin{figure}[h]
\centerline{\includegraphics[width=0.7\textwidth]{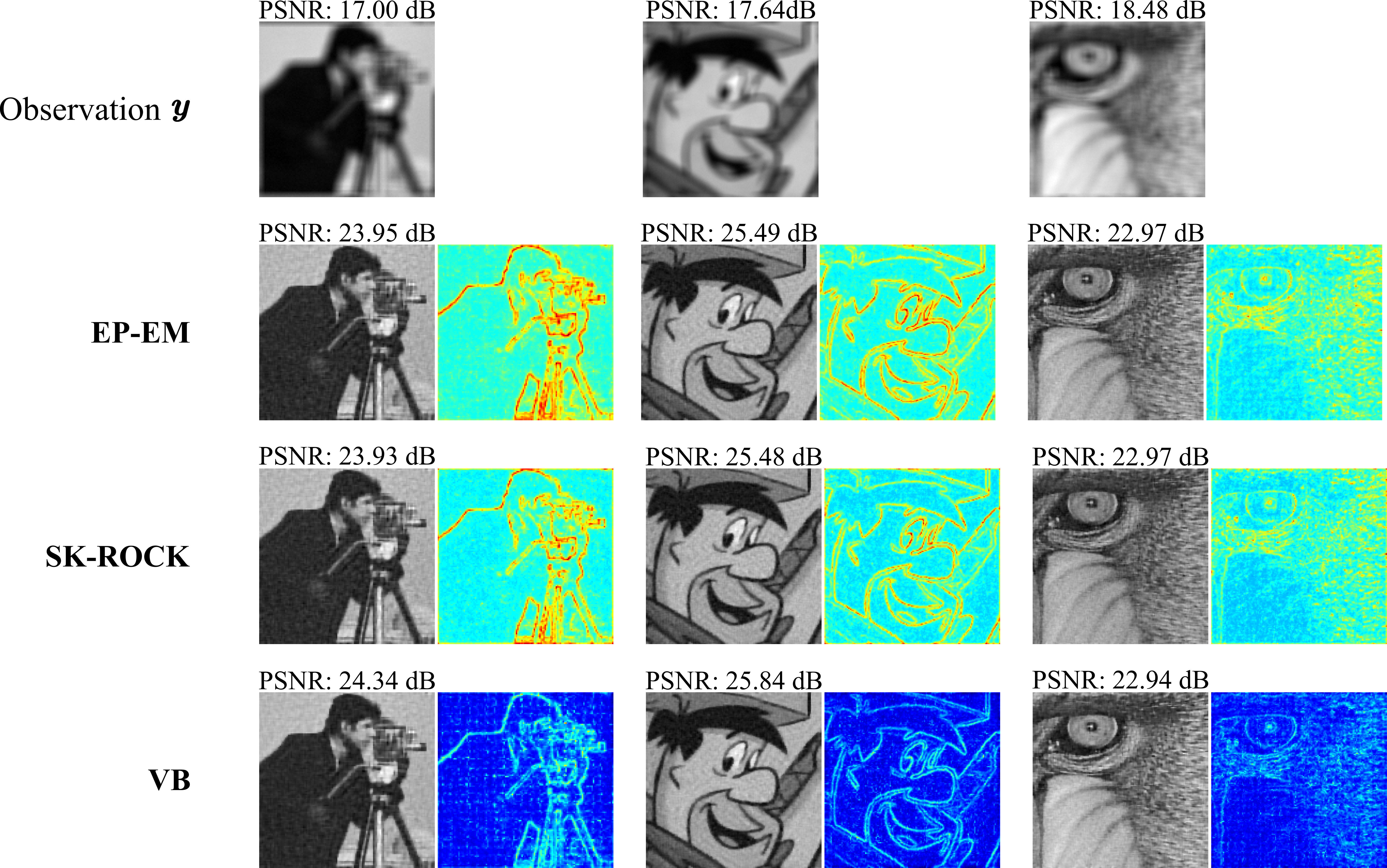}}
\caption{Comparison of deconvolution results obtained by EP-EM, SK-ROCK and VB, with $\ell_1$-TV prior using $\lambda_{\rm {EP-EM}}$.  The uncertainty estimates in the UQ maps are presented on a logarithmic scale.}
\label{fig: result_deconv_SKROCK_EP_VB}
\end{figure}

\begin{figure}[h]
\centerline{\includegraphics[width=0.7\textwidth]{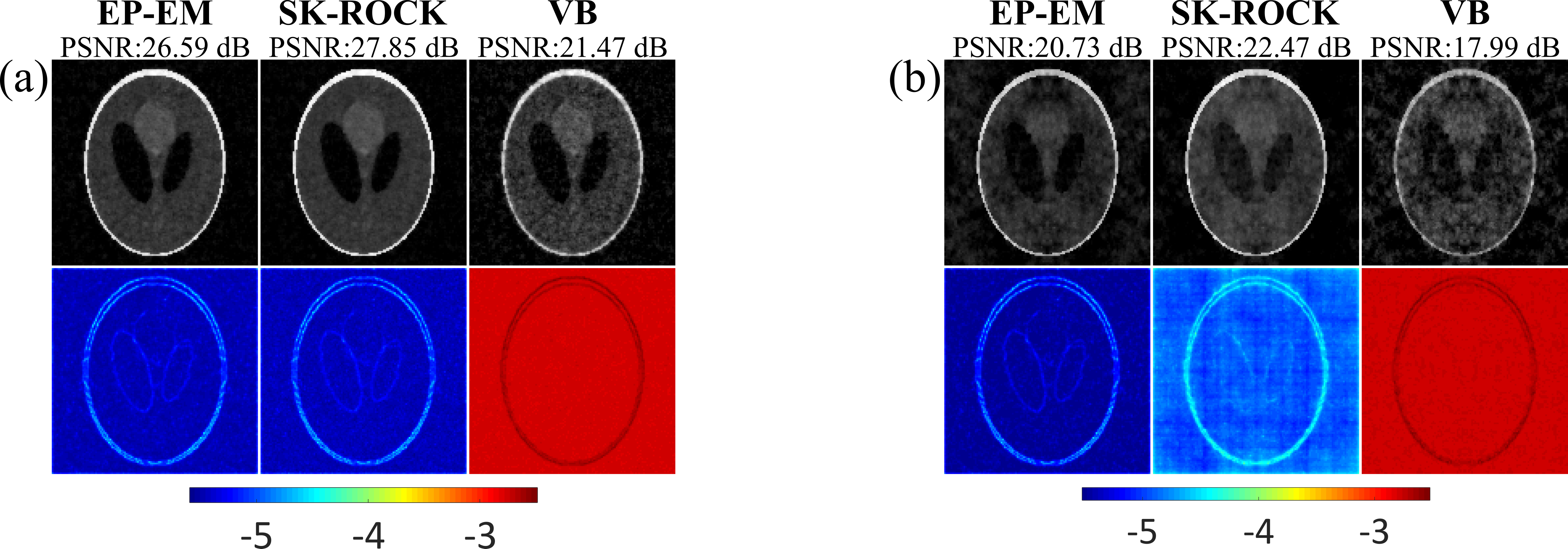}}
\caption{Comparison of CS reconstruction results obtained by EP-EM, SK-ROCK and VB, with $\ell_1$-TV prior using $\lambda_{\rm {EP-EM}}$. (a) $\textbf{H}$ is a Gaussian i.i.d. matrix with mean zero and variance $1/{M}$. (b) $\textbf{H}$ is a subsampled 2D Hadamard matrix. The uncertainty estimates in the UQ maps are presented on a logarithmic scale.}
\label{fig:result_CS_synthetic}
\end{figure}

We now investigate the quality of the approximate MMSE estimator and associated posterior marginal variances obtained via EP-EM. Fig. \ref{fig: result_deconv_SKROCK_EP_VB} shows the deconvolved  images and their UQ maps by EP-EM, SK-ROCK and VB, for $\textrm{BSNR} =35$ dB. Here again, SK-ROCK and VB are run with $\lambda_{\rm {EP-EM}}$. Compared to the denoising experiment in Fig. \ref{fig:result_denoising_UQ}, the three methods provide closer PSNRs, which seem to indicate smaller biases affecting the approximate MMSE estimators of EP and VB. Using SK-ROCK as reference, the EP bias is smaller than that of VB, whose mean tends to shift toward the mode of the posterior. The benefit of EP over VB in terms of UQ is more significant here, where EP slightly overestimates the marginal variances of SKROCK, while VB drastically underestimates the scale of the marginal variances.


\subsection{CS reconstruction}
We now evaluate the performance of 
EP-EM for CS reconstruction. Experiments are conducted on synthetic observations generated using \eqref{Eq: likelihood}, where ${\mathbf H}\in {\mathbb R}^{M\times N}$ is  a (a) Gaussian i.i.d. matrix with mean zero and variance $1/{M}$, and (b) 2D Hadamard matrix with $M$ randomly and uniformly selected patterns, $\xi =10^{-4}$. The reference $\boldsymbol x$ is the Shepp-Logan phantom image of size $128\times$128 pixels with pixel intensity in $[0,1]$.

Fig. \ref{fig:result_CS_synthetic} depicts
reconstruction results using EP-EM, SK-ROCK and VB using the same $\lambda_{\rm {EP-EM}}$ and $M/N=0.3$. In both cases, EP-EM provides an estimate $\lambda_{\rm {EP-EM}}$ which effectively regularizes the reconstruction problem. With $\lambda_{\rm {EP-EM}}$, EP and provides PSNRs closer to SK-ROCK than VB. While in (a) the UQ maps of EP and SK-ROCK are almost identical, EP underestimates, on average, the marginal uncertainties in (b), using SK-ROCK as reference. We believe this is due to the long-range pixel dependencies induced by the Hadamard patterns, and this observation will be further discussed in the conclusion of the paper. The marginal variances estimated by VB are significantly larger in both cases.




\subsection{Computational time}
All the experiments in this paper were carried out using MATLAB R2018b on an Intel(R) Core(TM) i7-8700K CPU @ 3.70GHz workstation. Table \ref{Table: computational_time} reports the computational time to obtain some of the results presented in Figs. \ref{fig:result_denoising_UQ},  \ref{fig: result_deconv_SKROCK_EP_VB}, and \ref{fig:result_CS_synthetic}.
The top row shows that the complexity of EP does not change significantly when changing the prior ($\ell_1$-TV slightly more expensive due to the truncated Gaussian distributions). While EP-EM would take approximately $T$ times longer than EP for $T$ EM iterations (with $\ell_1$-TV), its cheaper implementation, using a single update of the approximating factor, took only $5.5$ seconds (when EP takes $5.4$ seconds). The three bottom rows compare computational time of EP, SK-ROCK and VB, used to estimate posterior means and marginal variances (using the same $\lambda_{\rm EP-EM}$). EP is slightly faster than VB because it does not use large matrix multiplications during the approximation of the prior and the two variational methods are significantly faster than the sampling method, which requires many samples for accurate variance estimation. Note that SK-ROCK is less expensive in the deconvolution case mostly because the sub-routine used to compute proximity operators converges more quickly. The fast implementation of EP-EM in these cases has approximately the same cost as EP. If the estimation of $\lambda$ is the main objective, the cost of EP-EM is generally higher than competing methods (unless ${\mathbf H}^T{\mathbf H}$ is diagonal), as EB only requires a few iterations to estimate $\lambda$. However EP-EM remains attractive as it allows joint estimation of the image and $\lambda$, using a single fast algorithm, requiring minimum parameter tuning.

 \begin{table}[h]
\centering
\resizebox{0.5\textwidth}{!}{
\begin{tabular}{lcccc}
\toprule
\multirow{3}{*}{Fig. \ref{fig:result_denoising_UQ}}   & EP MoG2-TV & EP BG-TV & EP $\ell_1$-TV\\
& ($512\times 512$)    & (512$\times $512)  & ($512\times512$)   \\
\cmidrule(l){2-4} 
         & 4 seconds       & 4 seconds  & 5.4 seconds \\
\cmidrule(l){1-4}  
  & \multicolumn{3}{c}{$\ell_1$-TV ($128\times128$)}  \\
&EP & SK-ROCK ($\lambda_{\rm {EP-EM}}$)& VB ($\lambda_{\rm {EP-EM}}$)\\
\cmidrule(l){2-4} 
Fig. \ref{fig:result_denoising_UQ} &0.3 seconds  &14.3 hours &25 seconds  \\
Fig. \ref{fig: result_deconv_SKROCK_EP_VB} (a)   &40 seconds & 1 hour & 42 seconds \\
Fig. \ref{fig:result_CS_synthetic} (a) &15.9 minutes  &13 hours &23.5 minutes \\
\bottomrule 
\end{tabular}}
\caption{CPU computational time.}
\label{Table: computational_time}
\end{table}

\section{Conclusions and Future Work}
\label{Sec: Conclusion}

In this paper, we proposed a series of new EP algorithms with convex or non-convex gradient-based priors for scalable image restoration. We also discussed how EP can be embedded within more complex problems where additional problem parameters are unknown and can be estimated with EM-like procedures. We showed that the results are generally more accurate than using the VB alternative, and close to the MCMC-based alternative (which is taken as reference), at a fraction of the computational cost. 
In a denoising context, we showed that our fast EP methods can be used for rapid approximate MMSE estimation, together with marginal variance estimation. The method can be very easily adapted to Gaussian noise models with non-isotropic covariance matrices. Thus, it makes it particularly attractive for use within Plug-and-Play methods, beyond MAP-like denoisers.


The scalability of the proposed algorithms relies strongly on the diagonal structure of the covariance matrix of the global EP approximation. These constraints are well suited for most problems considered in this paper, where the exact posterior covariance matrix is close to diagonal. Indeed, the gradient-based priors do not induce strong long-range dependencies, nor does the matrix ${\mathbf H}$ if the noise level is sufficiently high. In such cases, the EP approximation is particularly accurate. However, the method is expected to fail if the posterior distribution exhibits strong correlation structures. In such cases, less restrictive covariance constraints should be considered, but how to keep the resulting variational method scalable would require further investigation.

\ifCLASSOPTIONcaptionsoff
  \newpage
\fi

\bibliographystyle{IEEEtran}
\bibliography{biblio.bib}

\end{document}